# A Novel Versatile Architecture for Autonomous Underwater Vehicle's Motion Planning and Task Assignment

Somaiyeh Mahmoud.Zadeh, David M.W Powers, Karl Sammut, Amir Mehdi Yazdani

**Abstract** Expansion of today's underwater scenarios and missions necessitates the requestion for robust decision making of the Autonomous Underwater Vehicle (AUV); hence, design an efficient decision making framework is essential for maximizing the mission productivity in a restricted time. This paper focuses on developing a deliberative conflict-free-task assignment architecture encompassing a Global Route Planner (GRP) and a Local Path Planner (LPP) to provide consistent motion planning encountering both environmental dynamic changes and *a priori* knowledge of the terrain, so that the AUV is reactively guided to the target of interest in the context of an uncertain underwater environment. The architecture involves three main modules: The GRP module at the top level deals with the task priority assignment, mission time management, and determination of a feasible route between start and destination point in a large scale environment. The LPP module at the lower level deals with safety considerations and generates collision-free optimal trajectory between each specific pair of waypoints listed in obtained global route. Re-planning module tends to promote robustness and reactive ability of the AUV with respect to the environmental changes. The experimental results for different simulated missions, demonstrate the inherent robustness and drastic efficiency of the proposed scheme in enhancement of the vehicles autonomy in terms of mission productivity, mission time management, and vehicle safety.

## 1 Introduction

Autonomous Underwater Vehicles (AUVs) generally are capable of spending long periods of time carrying out underwater missions at lower costs comparing manned vessels (Blidberg et al. 2001). An AUV needs to have a certain degree of autonomy to carry out mission objectives successfully and ensure safety in all stages of the mission, as failure is not acceptable due to expensive maintenance. Autonomous operation of AUV in a vast, unfamiliar and dynamic underwater environment is a complicated process, specifically when the AUV is obligated to react promptly to environmental changes. On the other hand, diversity of underwater scenarios and missions necessitates the requestion for robust decision making based on proper awareness of the situation. Hence, an advanced degree of autonomy at the same level as human operator is an essential prerequisite to trade-off within the problem constraints and mission productivity while manage the risks and available time. At the lower level, it again must autonomously carry out the collision avoidance and similar challenges. To this end, a hybrid architecture encompassing a Global Route Planner (GRP) and Local Path Planner (LPP), so that the AUV is reactively guided to the target of interest in the context of uncertain underwater environment. With respect to the combinatorial nature AUV's routing and task allocation, which is analogous to both Traveling Salesman Problem (TSP) and Knapsack problem, there should be a trade-off within the maximizing number of highest priority tasks with minimum risk percentage in a restricted time and guaranteeing reaching to the predefined destination, which is combination of a discrete and a continuous optimization problem at the same time. To provide a higher level of autonomy, deliberative hybrid architecture has been developed to promote vehicles capabilities in decision-making and situational awareness. To this end, the GRP module at the top level, simultaneously tends to determine the optimum route in terrain network cluttered with several waypoints and prioritize the available tasks. The proposed time optimum global route may have several alternatives, each of which consists of the proper sequence of tasks and waypoints. Another important issue that should be taken into consideration at all stages of the mission is vehicles safety, which is extremely critical and complicated issue in a vast and uncertain environment. The LPP module at lower level tends to generate the safe collision-free optimum path between pairs of waypoints included in the global route encountering dynamicity of the terrain; hence the LPP operates in context of the GRP module. Traversing the distance between two specific waypoints may take more time than expected due to dealing with dynamic unexpected changes of the environment.

S.M. Zadeh, D.M.W Powers, K.Sammut, A.M. Yazdani

School of Computer Science, Engineering and Mathematics, Flinders University,
Adelaide, SA, Australia
e-mail: somaiyeh.mahmoudzadeh@flinders.edu.au
e-mail: david.powers@flinders.edu.au
e-mail: karl.sammut@flinders.edu.au
e-mail: amirmehdi.yazdani@flinders.edu.au

S.M. Zadeh
School of Computer Science, Engineering and Mathematics, Flinders University,
Adelaide, SA, Australia

The loss of time in dealing with associated problem leads to requirement for a proper re-planning scheme. Hence, a "Synchro-module" is provided to manage the lost time within the LPP process and improve the robustness and reactive ability of the AUV with respect to the environmental changes. A variety of investigations have been carried out on autonomous unmanned vehicle motion planning and task allocation discussed in the next section.

The paper is organized in the following subsections. The related works to this research is provided in section 2. In section 3, the problem is formulated formally. An overview of the genetic algorithm and global route planner paradigm are presented in section 4. The particle swarm optimization and the local path planner are introduced in section 5. The architectures evaluation is discussed in section 6. The discussion on simulation results is provided in Section 7. And, the section 8 concludes the paper.

## 2 Related Works

Majority of AUV's motion planning approaches are categorized into two groups that first group attempt to find a trajectory that allows an AUV to transit safely from one location to another, while second group mostly concentrate on task allocation and vehicles routing problem (VRP). Respectively, the previous attempts in this scope are divided into two main categories as follows.

2.1 Vehicle Task Assignment-Routing

Effective routing has a great impact on vehicle time management as well as mission performance due to take selection and proper arrangement of the tasks sequence. Various attempts have been carried out in scope of single or multiple vehicle routing and task assignment based on different strategies. Karimanzira et al.(2014) presented a behaviour based controller coupled with waypoint tracking scheme for an AUV guidance in large scale underwater environment. Iori and Ledesma (2015) modelled AUVs routing problem with a Double Traveling Salesman Problem with Multiple Stacks (DTSPMS) for a single-vehicle pickup-and-delivery problem by minimizing the total routing cost. Other methods also studied on efficient task assignment for single/multiple vehicle moving toward the destination such as graph matching (Kwok et al. 2002), Tabu search (Higgins 2005), partitioning (Liu and Shell 2012), simulated annealing (Chiang and Russell 1996), branch and cut (Lysgaard et al. 2004), and evolutionary algorithms (Gehring and Homberger 2001). Martinhon et al.(2004) proposed stronger K-tree approach for the vehicle routing problem. Zhu and Yang (2010) applied an improved SOM-based approach for multi-robots dynamic task assignment. Alvarez et al.(2004) outlined a discrete method to grid the search space, then applied Genetic Algorithm (GA) on the grids to generate an energy optimal route. Also some popular graph search algorithms like A* (Al-Hasan and Vachtsevanos 2002; Pereira et al 2013) and Dijkstra (Eichhorn 2015) have been applied to determine a grid or cell-based route from the start to the destination point. Liu and Bucknall (2015) proposed a three-layer structure to facilitate multiple unmanned surface vehicles to accomplish task management and formation path planning in a maritime environment, in which the mission is divided between vehicles according to general mission requirement. Eichhorn (2015) implemented graph-based methods for the AUV ''SLOCUM Glider'' motion planning in a dynamic environment. The author employed modified Dijkstra Algorithm where the applied modification and conducted time variant cost function simplifies the determination of a time-optimal trajectory in the geometrical graph. Wang et al. (2005) introduced an adaptive genetic algorithm to determine real-time obstacle-free route for AUV in a large-scale terrain in presence of few waypoints. An energy efficient fuzzy based route planning using priori known wind information in a graph-like terrain is presented by Kladis et al. (2011) for UAVs motion planning. M.Zadeh et al. (2015) investigated a large-scale AUV routing and task assignment joint problem by transforming the problem space into a NP-hard graph context, in which the heuristic search nature of GA and PSO employed to find the best series of waypoints. This work is extended to semi dynamic networks while two biogeography-based optimization (BBO) and PSO meta-heuristic algorithms are adopted to provide real-time optimal solutions (M.Zadeh et al. 2016-a). The traditional algorithms used for graph routing problem have major shortcomings (notably high computational complexity) for real-time applications. Majority of the discussed research, in particular, focus on task and target assignment and time scheduling problems without considering requirements for vehicle safe deployment toward the destination.

2.2 Path/Trajectory Planning Approaches

Many strategies have been provided and applied to the AUV path-planning problem in recent years encountering dynamicity of the terrain. Methods like D* or A* algorithms have been employed for AUV optimum path generation (Carsten et al. 2006; Likhachev et al.2005). Another approach to solve this problem is the Fast Marching (FM) algorithm, which uses a first order numerical approximation of the nonlinear Eikonal equation. Petres et al. (2005) provided FM-based path planning to deal with a dynamic environment. This method is accurate but also computationally expensive than A*. Later on, an upgraded version of FM known as FM* or heuristically guided FM is investigated on path planning problem (Petres et al. 2007) that preserves the efficiency of the FM and accuracy of the A* algorithm, while apparently it is restricted to use linear anisotropic cost to attain computational efficiency. In particular, the main drawback of these methods is that their time complexity increases exponentially with increasing the problem space. Generally, the heuristic grid-search based methods are criticized because their discrete state transitions, which restrict the vehicle's motion to limited directions.

Another solution for path planning is using the evolution based algorithms. Evolutionary algorithms are population based optimization methods that can be implemented on a parallel machine with multiple processors to speed up computation (Roberge et al. 2013). Relatively, they are efficient methods for dealing with path planning as a Non-deterministic Polynomial-time (NP) hard problem (M.Zadeh et al. 2016-c; Ataei and Yousefi-Koma 2015), and fast enough to satisfy the time restrictions of real-time applications. The Particle Swarm Optimization (Zheng et al. 2005) and Genetic Algorithm (Nikolos et al.2003; Zheng et al. 2005; Kumar and Kumar 2010) are two popular types of optimization algorithms applied successfully in path planning application. Fu et al. (2012) employed Quantum-based PSO (QPSO) for unmanned aerial vehicle path planning, but implemented only off-line path planning in a static and known environment, which is far from reality. Subsequently, this algorithm was employed by Zeng et al., (2014-a; 2014-b) for on-line AUV path planning in a dynamic marine environment. A Differential Evolution based path planning is proposed by (M.Zadeh et al. 2016-b) for the AUV operation in three-dimensional complex turbulent realistic underwater environment.

Although various path-planning techniques have been suggested for autonomous vehicles, AUV-oriented applications still have several difficulties when operating across a large-scale geographical area. The computational complexity grows exponentially with enlargement of search space dimensions. To cope with this difficulty, speed up the path planning process and reduce memory requirement, the majority of conventional path planning approaches transmuted the 3D environment to 2D space. However, a 2D representation of a marine environment doesn't sufficiently embody all the information of a 3D ocean environment and vehicle motion with six degrees of freedom. In large-scale operations it is hard to estimate all probable changes of the terrain (obstacles/current behavior) and tracking the behavior of a dynamic terrain beyond the vehicles sensor coverage is impractical and unreliable. A further problem is then; a huge amount of data about the update of entire terrain condition must be computed repeatedly. This huge data load from environment should be analyzed continuously every time path replanting is required, which is computationally inefficient and unnecessary as only awareness of environment in vicinity of the vehicle such that the vehicle can be able to perform reaction to environmental changes is enough. On the other hand, when the terrain is cluttered with multiple waypoints and the vehicle is requested to carry out a specific sequence of prioritized tasks assigned to trajectories between waypoints, path planning is not able to facilitate the vehicle to carry out the task assignment considering graph routing restrictions; thus, a routing strategy is required to handle graph search constraints and facilitate the task assignment. The routing strategies are not as flexible as path planning in terms coping with environmental prompt changes, but they give a general overview of the area that AUV should fly through (general route), which means cut off the operation area to smaller beneficial zones for deployment. To summarize above discussion, existing approaches mainly cover only a part of the AUV routing task assignment problem, or path planning along with obstacle avoidance as a safety consideration.

2.3  Research Contribution

To carry out the underwater missions in large scale environment in presence of severe environmental disturbances, a hybrid architecture with re-planning capability is being developed to cover shortcomings and to take advantages of both path and route planning strategies, which is a significant change to accelerate the computational runtime. The proposed system is designed in separate modules running concurrently including a global route planner (GRP) at top level with higher level of decision autonomy, and the local path planner (LPP) at lower level, to autonomously carry out the collision avoidance. A constant interaction is flowing between these two modules by back feeding the situational awareness of the surrounding operating filed form the LPP to the GRP for making a decision on requisitions of re-planning. Hence, the third module "Synchro-module" is performed to manage the lost time within the LPP/GRP process and reactively adapt the system to the last update of environment and decision parameters (e.g. remaining time).This process continues iteratively until the AUV reaches the end point. A significant benefit of such detached design is that different methods employed by main modules and even sub modules can be easily replaced with new methods or get upgraded without requiring any change in whole structure of the system. This issue specifically increases the reusability of the control architecture and specifically eases updating/upgrading AUV's maneuverability at all times.

Obtaining the exact optimum solution is only possible for the specific case where the environment is completely known and no uncertainty exists and the environment modelled by this research corresponds to a dynamic environment with high uncertainty. Moreover, the task and route planning problem is a generalization of both the knapsack and TSP problems and meta–heuristics are the fastest approach introduced for solving NP-hard complexity of these problems and have been shown to produce solutions close to the optimum with high probability (Iori and Ledesma, 2015; Besada-Portas, et al., 2010). On the other hand, precise and concurrent synchronization of the higher and lower level modules is the primary requirement for preserving the consistency, stability and cohesion of this real-time system in meeting the specified objectives. The most critical factor for both GRP and LPP operation is having a short computational time to provide fast concurrent synchronization between modules while balancing the constraints. Maintaining comparably fast operation for each component of the main architecture is necessary to prevent any of them from dropping behind the others. Any such a delay disrupts the routine flow and concurrency of the entire system, and adding NP computational time into the equation would itself render a solution suboptimal. While the solutions proposed by any meta–heuristic algorithm do not necessarily correspond to the optimal solution, it is more important to control the time, and thus we rely on the previously mentioned ability of meta-heuristic algorithms, including GA and PSO algorithms as employed in the GRP and LPP modules, to find correct and near optimal solutions in competitive time (real and CPU).

## 3 Problem Formulation

The main goal of AUV operation is to complete mission objectives while ensuring the vehicle's safety at all times. Appropriate vehicle routing and path planning strategy along with efficient synchronization between models maximizes the achievements of a mission. A mathematical formulation of the problem is provided in the following subsections.

### 3.1 Mathematical Representation of the Operation Terrain

The ocean environment is modelled as a three-dimensional terrain $\Gamma_{3D}$ covered by uncertain, static and moving obstacles comprising several fixed waypoints. An underwater mission is commenced at a specified starting point $WP^1:(x_1,y_1,z_1)$ and it is terminated when the AUV reaches to a predefined destination point (dock for example) at $WP^D:(x_D,y_D,z_D)$. The waypoints' location are randomized according to a uniform distribution of $\sim U(0,10000)$ for $WP^i_{x,y}$ and $\sim U(0,100)$ for $WP^i_z$. Waypoints in the terrain are connected with edge like $q_i$ from a set of $q=\{q_1,...,q_m\}$, where $m$ is the number of edges in the graph.

Each edge of the network like $q_i$ is assigned with a specific task from a set of Task=$\{Task_1,...,Task_k\}$ $k \in m$ in advance. Each task has a value like $\rho_i$ from a limited set of $\rho=\{\rho_1,...,\rho_k\}$ that represents its priority comparing other tasks, and completion time of $\delta_T$ regardless of required time for passing the relevant edge. Each task also has a risk percentage of $\xi_T$ regardless of terrain hazards and risks. Exploiting a priori knowledge of the underwater terrain, the initial step is to transform the problem space into a graph problem as depicted in *Fig.1*; then the GRP module tends to find the best fitted route to the available time, involving the best sequence of waypoints.

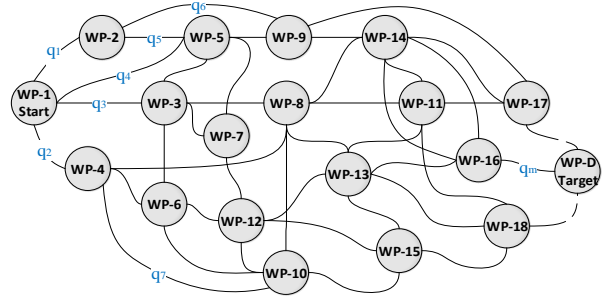

**Fig. 1.** A graph representation of operating area covered by waypoints

$$q_{ij}:\begin{cases}d_{ij}\\t_{ij}\end{cases} \qquad Task_{q_{ij}}:\begin{cases}\rho_{T_{ij}}\\\delta_{T_{ij}}\\\xi_{T_{ij}}\end{cases} \qquad (1)$$

The underwater variable environment poses several challenges for AUV deployment, such as dealing with static, dynamic and uncertain obstacles and ocean current. Ocean current is considered as static current that affect floating obstacles.

### 3.2 Mathematical Representation of Static/Dynamic Obstacle

In terms of collision avoidance, obstacle's velocity vectors and coordinates can be measured by the sonar sensors with a certain uncertainty modelled with a Gaussian distributions. The state of obstacle(s) continuously measured and sent to state predictor to provide the estimation of the future states of the obstacles for the LPP. The state predictor estimates the obstacles behaviour during the vehicles deployment in specified operation window. Four different type of obstacles are conducted in this study to evaluate the performance of the proposed path planner, in which an obstacle is presented by three components of position, radius and uncertainty $\Theta_{(i)}:(\Theta_p,\Theta_r,\Theta_{Ur})$. The obstacle position $\Theta_p$ initialized using normal distribution of $\sim\mathcal{N}(0,\sigma^2)$ bounded to position of start waypoint $WP^a_{x,y,z}$ and position of target waypoint $WP^b_{x,y,z}$, where $\sigma^2 \approx \Theta_r$. Therefore the obstacles position $\Theta_p$ on $WP^a_{x,y,z} < \Theta_p < WP^b_{x,y,z}$ has a truncated normal distribution, where its probability density function defined as follows:

$$\Theta^i_p \in \left(WP^a_{x,y,z}, WP^b_{x,y,z}\right) - \Theta^i_r \qquad (2)$$

$$f(\Theta^i_p; 0, \Theta^i_r, WP^a_{x,y,z}, WP^b_{x,y,z}) = \frac{\Theta^i_p}{(\Theta^i_r)^2} \Bigg/ \left(\frac{WP^b_{x,y,z} - \Theta^i_p}{\Theta^i_r} - \frac{WP^a_{x,y,z} - \Theta^i_p}{\Theta^i_r}\right) \qquad (3)$$

The obstacle radius initialized using a Gaussian normal distribution of $\sim(0,100)$. Different type of considered obstacles in this research explained in the rest.

- **[1] Static Known Obstacles:** The location of these obstacles is known in advance and their position can be obtained from offline map. No uncertainty growth considered for position of these obstacles (e.g. known rocks in the terrain).
- **[2] Static Obstacles with Certain Growth of Uncertainty**: These obstacles classified as the Quasi-static obstacles that usually known as no flying zones. The obstacles in this category has an uncertain radius varied in a specified bound with a Gaussian normal distribution $\sim(\Theta_p,\sigma_0)$, where the value of $\Theta_r$ in each iteration is independent of its previous value.
- **[3] Self-Motivated Moving Obstacle:** Self-motivated moving obstacle, is the third type that has a motivated velocity that shift it from position A to position B. Therefore, its position changes to a random direction with an uncertainty rate proportional to time, given in (4).
  $$\Theta_p(t) = \Theta_p(t-1) \pm U(\Theta_{p_0}, \sigma) \qquad (4)$$
- **[4] Moving Obstacles with Propagated Uncertainty in Position and Radius:** Another type of considered obstacle is self-motivated moving obstacle that affected by current force and moving with a self-motivated velocity to a random direction, denoted by (4) and (5). Here, the effect of current presented by uncertainty propagation proportional to current magnitude $U_R^C = |V_C| \sim (0,0.3)$ that radiating out from the center of the obstacle in a circular format.

$$\Theta_r(t) = B_1\Theta_r(t-1) + B_2 X_{(t-1)} + B_3\Theta_{Ur}$$
$$B_1 = \begin{bmatrix} 1 & U_R^C(t) & 0 \\ 0 & 1 & 0 \\ 0 & 0 & 1 \end{bmatrix}, \quad B_2 = \begin{bmatrix} 0 \\ 1 \\ 1 \end{bmatrix}, \quad B_3 = \begin{bmatrix} 0 \\ 0 \\ U_R^C(t) \end{bmatrix} \quad (5)$$

where $\Theta_{Ur} \sim \sigma$ is the rate of change in objects position, and $X_{(t-1)} \sim \mathcal{N}(0,\sigma_0)$ is the Gaussian normal distribution that assigned to each obstacle and gets updated in each iteration $t$.

### 3.3 Mathematical Representation of the AUV Routing Problem

The mathematical representation of the AUVs' routing problem should be simple enough to avoid unnecessarily expensive computations. The generated route should be applicable and logically feasibleaccording to feasibility criteria's (given in section 4), and represented by $R_i=(x_1,y_1,z_1,...,x_i,y_i,z_i,...,x_D,y_D,z_D)$, where $(x_i,y_i,z_i)$ is the coordinate of any arbitrary waypoint in geographical frame. The goal is to find the optimum route covering the maximum number of highest priority tasks with smallest risk percentage in a time interval that battery's capacity allows. The problem involves multiple objectives that should be satisfied during the optimization process. In the preceding discussion, the mathematical representation of the AUV route planning problem in $\Gamma_{3D}$ terrain is described as follows:

$$d_{ij} = \sqrt{(x_j - x_i)^2 + (y_j - y_i)^2 + (z_j - z_i)^2} \quad (6)$$

$$t_{ij} = \frac{d_{ij}}{V_{AUV}} + \delta_{Tij} \quad (7)$$

$$T_{Route} = \sum_{\substack{i=0 \\ j \neq i}}^{n} lq_{ij}t_{ij} = \sum_{\substack{i=0 \\ j \neq i}}^{n} lq_{ij}\left(\frac{d_{ij}}{V_{AUV}} + \delta_{Tij}\right), \quad l \in \{0,1\} \quad (8)$$

$$W_R = \sum_{\substack{i=0 \\ j \neq i}}^{n} lq_{ij}\frac{\rho_{ij}}{\xi_{ij}} = \sum_{\substack{i=0 \\ j \neq i}}^{n} lq_{ij}w_{ij} \Rightarrow \max\left(\sum_{\substack{i=0 \\ j \neq i}}^{n} lq_{ij}w_{ij}\right), \quad l \in \{0,1\} \quad (9)$$

$$\min\left(|T_{Route} - T_{Available}|\right) = \min\left(\left|\sum_{\substack{i=0 \\ j \neq i}}^{n} lq_{ij}\left(\frac{d_{ij}}{V_{AUV}} + \delta_{Tij}\right) - T_{Available}\right|\right) \quad (10)$$

s.t.
$$\max(T_{Route}) < T_{Available}$$

where $T_{Route}$ is the required time to pass the route, $T_{Available}$ is the total mission time, $l$ is the selection variable, $t_{ij}$ is the required time to pass the distance $d_{ij}$ between two waypoint of $WP^i$ and $WP^j$ along with task completion time $\delta_{Tij}$. $\rho_{Tij}$ and $\xi_{Tij}$ are the priority value and risk percentage of the task, respectively. The next step is generating time/distance optimum trajectory in smaller scale between each pair of waypoints in optimum global route.

### 3.4 Path Planning Problem Formulation

The path planner should generate time optimum collision-free local path $\wp_i$ (shortest path) between specific pair of waypoints through a spatiotemporal underwater environment in the presence different types of uncertain obstacles. The resultant path should be safe and flyable (feasible). The operation terrain modelled as a time varying environment covered by uncertain, static and moving obstacles $\Theta$ mentioned above. The dimension of the operating window depends on distance between two nominated waypoints. The proposed path planner in this study, generates the potential trajectories using B-Spline curves captured from a set of control points like $\vartheta = \{\vartheta_1, \vartheta_2, ..., \vartheta_i, ..., \vartheta_n\}$ in the problem space with coordinates of $\vartheta_1:(x_1,y_1,z_1),...,\vartheta_n:(x_n,y_n,z_n)$, where $n$ is the number of corresponding control points. These control points play a substantial role in determining the optimal path. The mathematical description of the B-Spline coordinates is given by:

$$X(t) = \sum_{i=1}^{n} x_i B_{i,K}(t)$$
$$Y(t) = \sum_{i=1}^{n} y_i B_{i,K}(t) \quad (11)$$
$$Z(t) = \sum_{i=1}^{n} z_i B_{i,K}(t)$$

where $B_{i,K}(t)$ is the curve's blending functions, $t$ is the time step, and $K$ is the order of the curve and shows the smoothness of the curve, where bigger $K$ correspond to smoother curves represented in *Fig*.2. For further information refer to [24].

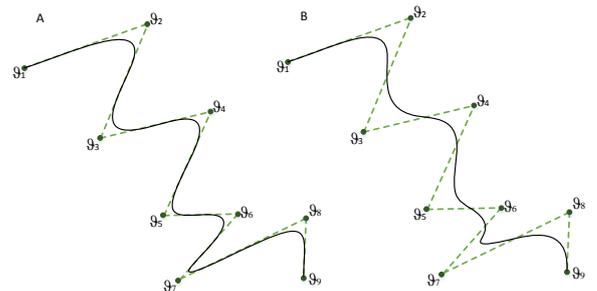

**Fig. 2.** Quadratic B-Spline curve by control points, where in (A), K = 3.5 and in (B), K = 6.

The path-travelled time $T_{path\text{-}flight}$ between two waypoints for $\wp_i$ should be minimized. The ocean current velocity is assumed to be constant. The path planner is applied in a small-scale area, and the AUV considered to have constant thrust power; therefore, the battery usage for a path is a constant multiple of the time and distance travelled. Performance of the generated trajectory is evaluated based on overall collision avoidance capability and time consumption, which is proportional to energy consumption and travelled distance. The path planner's cost function is detailed in section 6.2.

To cope with the probable challenges of the dynamic environment, the LPP repeatedly calculates the trajectory between vehicles current position and its specified target location. The path absolute time $t_{ij}$ is calculated at the end of the trajectory. Then $t_{ij}$ is added to corresponding task completion time $\delta_{Task}$ and the computation time $T_{compute}$. Total value of this summation $T_{path\text{-}flight}$ gets compared to expected time $T_{Expected}$ for passing the distance between specified pair of waypoints. If $T_{path\text{-}flight}$ gets smaller value than $T_{Expected}$, it means no unexpected difficulty is occurred and vehicle can continue its travel along the current global route. However, if $T_{path\text{-}flight}$ exceeds the $T_{Expected}$, it means AUV faced a challenge during its deployment. Obviously, a specific amount of battery and time $T_{Avaliable}$ is wasted for handling collision avoidance, so the $T_{Avaliable}$ should be updated. In such case, the current route cannot be optimum anymore due to loss of time and re-planning is required according to mission updates.

$$T_{path\text{-}flight} = t_{ij} + \delta_{Task_{ij}} + \sum T_{compute} \qquad (12)$$

> **if** $T_{path\text{-}flight} \leq T_{Expected}$
>   Continue the current optimum route $R_j$
> **else if** $T_{path\text{-}flight} > T_{Expected}$
>   Update $T_{Avaliable}$ and operation network
>   Re-plan a new route according to mission updates

It would be computation and time dissipation for an AUV to pass a specific edge (distance) for more than ones that means repeating a task for several times. Hence, if re-routing is required at any situation, the $T_{Avaliable}$ gets updated; the passed edges get eliminated from the operation network (so the search space shrinks); and the location of the present waypoint is considered as the new start position for both LPP and GRP. Afterward, the GRP tends to find the optimum route based on new information and updated network topology. The process of combinational GRP, LPP, re-planning process and schematic representation of proposed control architecture is provided in *Fig*.3 and *Fig*.4, respectively.

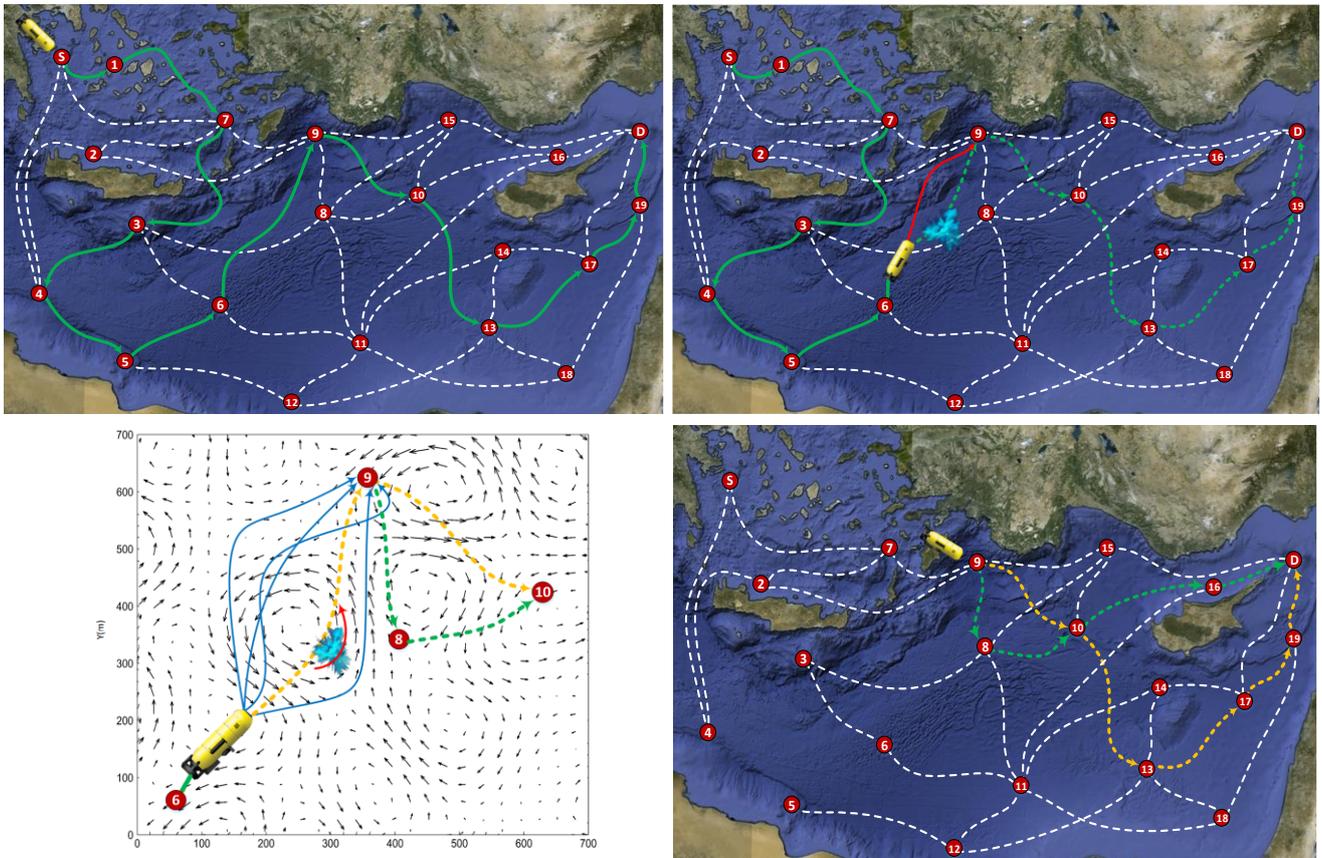

**Fig.3.** Graph representation of operating area covered by waypoints and route/ trajectory planning, re-planning process

Given a candidate route in a sequence of waypoints (e.g. initial optimum route: {S-1-7-3-4-5-6-9-10-13-17-19-D} in *Fig*.3) along with environment information, the LPP module provides a trajectory to safely guide the vehicle through the waypoints. During deployment between two waypoints, the LPP can incorporate any dynamic changes of the environment. The provided trajectory is then sent to the guidance controller to generate the guidance commands for the vehicle. After visiting each waypoint, the re-planning criteria (given in equation (12)) is investigated. If re-planning is required, the "Synchro-module" updates the operation graph and mission available time; and the controller recalls the GRP to provide new optimum route based on mission updates(e.g. new optimum route: { 9-8-10-16-D}). This process continues until mission ends and vehicle reaches the required waypoint.

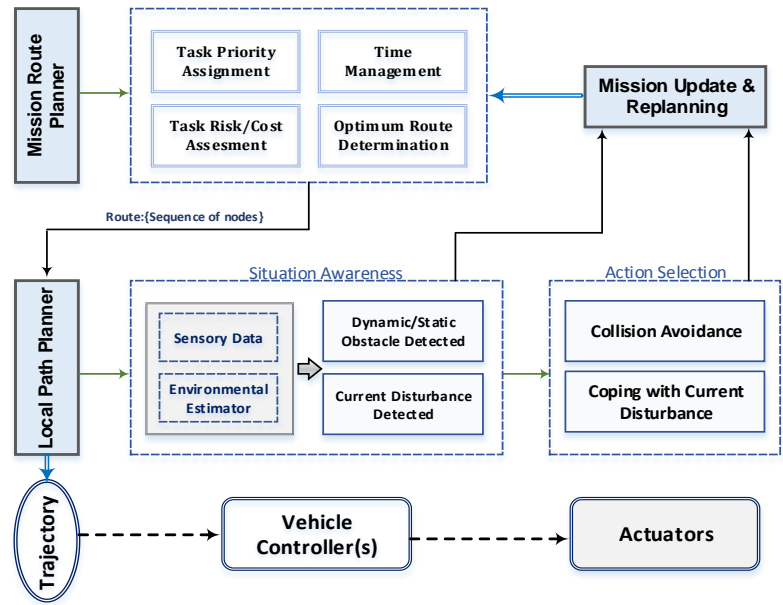

**Fig.4.**Proposed control architecture including AUV's cooperative GRP and LPP scheme

The trade-off between available mission time and mission objectives is critical issue that can be adaptability carried out by GRP. Hence, the main synchronous architecture should be fast enough to track environmental changes, cope with dynamic changes, and carry out prompt re-planning. To handle the complexity of NP-hard graph routing and task allocation problem, the GRP takes the advantages of genetic algorithm to find an optimum global route for the underwater mission. In the LPR module, the particle swarm optimization algorithm carries out path planning between each pair of the waypoints, which is efficient and fast enough in generating collision-free optimum trajectory in smaller scale.

## 4 Overview of Genetic Algorithm and Global Routing Process

The Genetic Algorithm (GA) is a particular type of stochastic optimization search algorithm represents problem solving technique based on biological evolution. GA has been extensively studied and widely used on many fields of engineering. It searches in a population space that each individual of this population is known as chromosome. Its process starts with randomly selecting a number of feasible solutions from the initial population. A fitness function should be defined to evaluate quality of solutions during the evolution process. New population is generated from initial population using the GA operators like selection, crossover and mutation. Chromosomes with the best fitness value are transferred to next generation and the rest get eliminated. This process continues until the chromosomes get the best fitted solution to the given problem (Sivanandam and Deepa 2008). The average fitness of the population gets improved at each iteration by adaptive heuristic search nature of the GA. The GRP module deals with finding the optimal route through the operating graph, where the input to this module is a group of feasible generated routes involving a sequence of nodes with same starting and ending points that are encoded as chromosomes. The operation is terminated when a fixed number of iterations get completed, or when no dramatic change observed in population evolution. The process of the GA algorithm is proposed by the flowchart given in *Fig*.5.

Developing a suitable coding scheme and chromosome representation is the most critical step of formulating the problem in GA framework. Hence, efficient representation of the routes and encoding them correctly into the chromosomes has direct impact on overall performance of the algorithm and optimality of the solutions. The process of The GA-based algorithm for route planning and task priority assignment is summarized in following steps.

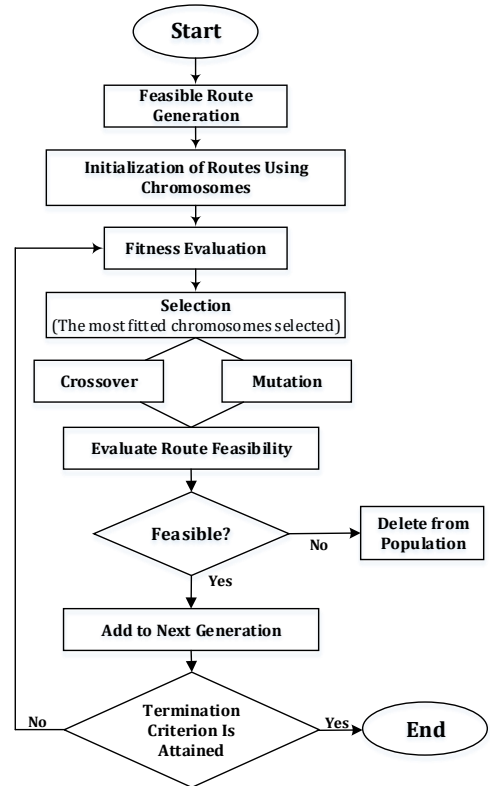

**Fig.5.** Process of the GA algorithm

### 4.1 *Chromosome Encoding (Initialize Chromosome/Route Population)*

A chromosome in the proposed GA corresponds to a feasible route including a sequence of nodes. The first and last gene of the chromosomes always corresponds to the start and destination node with respect to the topological information of the graph. Chromosomes take variable length, but limited to maximum number of nodes included in the graph, since it is never required for a route to include nodes more than whole number of nodes in the graph. The resultant solution from both GA should be feasible and valid according to criteria's given in *Fig*.6. A priority based strategy is used in this research to generate feasible routes [34]. For this purpose, some guiding information is added to each node at the initial phase. The priority vector initialized randomly. The nodes are selected based on their corresponding value in priority vector and Adjacency relations. Using Adjacency matrix prevents appearance of non-existed edges of the graph. To prevent generating infeasible routes some modifications are applied as follows:

- A valid route should be commenced and ended with predefined start and target nodes.
- The generated route should not include edges that are not presented in the graph.
- The multiple appearance of the same node in a route makes it invalid because this issue implies wasting time repeating a task.
- The route should not traverse an edge for more than once.
- The route travel time should not exceed the maximum range of total available time.

**Fig.6.** Route feasibility criteria

- Each node take positive or negative priority values in the specified range of [-100,100]. The selected node in a route sequence gets a large negative priority value that prevents repeated visits to a node. Then, the visited edges get eliminated from the Adjacency matrix. So that, the selected edge will not be a candidate for future selection. This issue reduces the memory usage and time complexity for large and complex graphs.

- To satisfy the termination criteria of a feasible route, if the route ends with a non-destination node and/or the length of the route exceeds the number of existed nodes in the graph, the last node of the sequence get replaced by index of the destination node. This process keeps the generated route in feasible (valid) space.

Figure 7 presents an example of the route generating process according to a sample Adjacency matrix(*Ad*) of a graph and a random priority array ($U_i$). To generate a feasible route in a graph with 18 nodes based on topological information, the first node is selected as the start position. Then from Adjacency matrix the connected nodes to node-1 are considered. In graph shown in *Fig*.7, this sequence is {*2,3,4,5*}. The node with the highest priority in this sequence is selected and added to the route sequence as the next visited node. This procedure continue until a legitimate route is built (destination visited).

| *Ad* | Example of adjacency matrix for a graph with 18 nodes |
| *n* | Node index where *n=1* is the start and *n=18* is the destination point |
| $R^k_{Ui}$ | Partial route corresponding to the priority vector of a route including *k* nodes. |
| $U_i$ | Priority array (random no repeated vector in range of [-100,100]) |

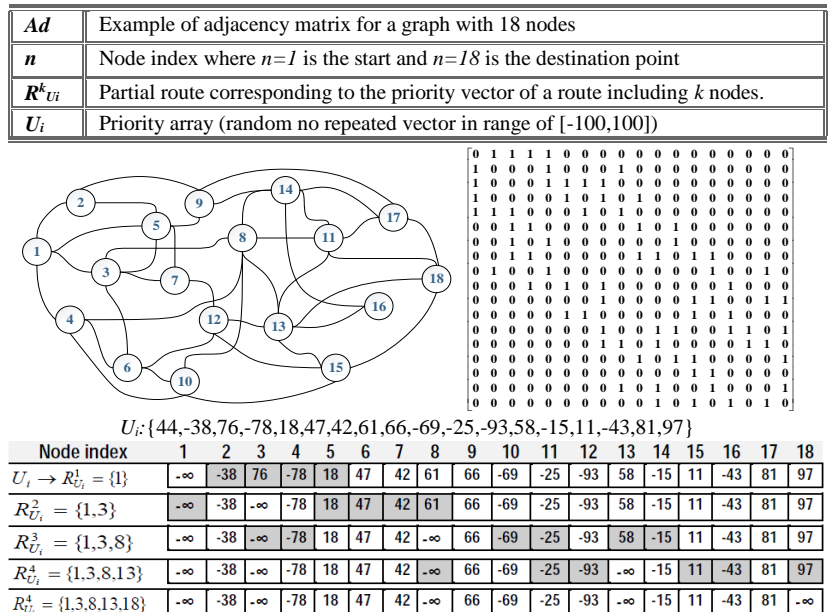

$U_i$:{44,-38,76,-78,18,47,42,61,66,-69,-25,-93,58,-15,11,-43,81,97}

| Node index | 1 | 2 | 3 | 4 | 5 | 6 | 7 | 8 | 9 | 10 | 11 | 12 | 13 | 14 | 15 | 16 | 17 | 18 |
|---|---|---|---|---|---|---|---|---|---|---|---|---|---|---|---|---|---|---|
| $U_i \to R^1_{U_i} = \{1\}$ | -∞ | -38 | 76 | -78 | 18 | 47 | 42 | 61 | 66 | -69 | -25 | -93 | 58 | -15 | 11 | -43 | 81 | 97 |
| $R^2_{U_i} = \{1,3\}$ | -∞ | -38 | -∞ | -78 | 18 | 47 | 42 | 61 | 66 | -69 | -25 | -93 | 58 | -15 | 11 | -43 | 81 | 97 |
| $R^3_{U_i} = \{1,3,8\}$ | -∞ | -38 | -∞ | -78 | 18 | 47 | 42 | -∞ | 66 | -69 | -25 | -93 | 58 | -15 | 11 | -43 | 81 | 97 |
| $R^4_{U_i} = \{1,3,8,13\}$ | -∞ | -38 | -∞ | -78 | 18 | 47 | 42 | -∞ | 66 | -69 | -25 | -93 | -∞ | -15 | 11 | -43 | 81 | 97 |
| $R^4_{U_i} = \{1,3,8,13,18\}$ | -∞ | -38 | -∞ | -78 | 18 | 47 | 42 | -∞ | 66 | -69 | -25 | -93 | -∞ | -15 | 11 | -43 | 81 | -∞ |

**Fig.7.** Sample of feasible route generation process based on topological information (priority vector $U_i$ and Adjacency matrix *Ad*) ref p1

### 4.2 Selection
Selecting the parents for crossover and mutation operations is another step of the GA algorithm that plays an important role in improving the average quality of the population in the next generation. Several selection methods exist for this purpose such as roulette wheel, ranks selection, elitist selection, scaling selection, tournament selection, etc. The roulette wheel selection has been conducted by current research, wherein the next generation is selected based on corresponding cost value, then the wheel divided into a number of slices and the chromosomes with the best cost take larger slice of the wheel.

### 4.3 Crossover Operation
Crossover is a GA operator that shuffles sub parts of two parent chromosomes and generate offspring that includes some part of both parent chromosomes. Many types of crossover techniques have been suggested since now. Generally, they can be categorized in to two main types of single point and multipoint crossover methods. In a single point crossover, only one crossing site exists, while in multipoint crossover, multiple sites of a pair of parents are selected randomly to get shuffled. The single point crossover method is simple, but it has some drawbacks like formation of loop (cycles) when applied for routing problem. Therefore, to prevent such an issue it is required to use more advanced type of multipoint crossover method like Order crossover (OX), Cycle crossover (CX), Partially Matched (PMX), Uniform crossover (UX) and so on (Sivanandam and Deepa 2008). Discussion over which crossover method is more appropriate still is an open area for research. Current research took advantages of uniform crossover, which uses a fixed mixing ratio among pair of parents. The gens are swapped with a fixed probability that usually is considered as 0.5. This method is extremely useful in problems with a very large search space in those where recombination order is important. An example of uniform crossover is given below:

| | | | | | | | | | | |
|---|---|---|---|---|---|---|---|---|---|---|
| *Parent-1:* | $WP_S$ | $WP_3$ | $WP_{14}$ | $WP_{18}$ | $WP_8$ | $WP_4$ | $WP_7$ | $WP_{17}$ | $WP_D$ | |
| *Parent-2:* | $WP_S$ | $WP_5$ | $WP_9$ | $WP_6$ | $WP_{11}$ | $WP_{16}$ | $WP_{13}$ | $WP_{10}$ | $WP_{12}$ | $WP_{19}$ | $WP_D$ |
| *Offspring-1:* | $WP_S$ | $WP_5$ | $WP_{14}$ | $WP_6$ | $WP_{11}$ | $WP_4$ | $WP_{13}$ | $WP_{10}$ | $WP_D$ | |
| *Offspring-2:* | $WP_S$ | $WP_3$ | $WP_9$ | $WP_{18}$ | $WP_8$ | $WP_{16}$ | $WP_7$ | $WP_{17}$ | $WP_{12}$ | $WP_{19}$ | $WP_D$ |

**Fig.8.** Example of uniform crossover

If the length of the chromosome is smaller than four, the crossover operator gets in to trouble of finding crossing site and swapping. So the chromosomes with length less than four gens get discarded from the crossover operation. The offspring gets eliminated if it does not correspond to a feasible route.

### 4.4 Mutation Operation

Mutation is another GA operator for generating the new population. This operator provides bit flipping, insertion, inversion, reciprocal exchange, etc., for altering parents (Sivanandam and Deepa 2008). Current research applies a combination of three inversion, insertion, and swapping type of mutation methods, explained in *Fig*.9. All these three methods preserve most adjacency information. In order to keep the new generation in feasible space, the mutation is applied on gens between the first and last gens of the parent chromosome that correspond to start and destination point. Both of the mutation and crossover operations enhance the rate of convergence.

| $WP_S$ | $WP_6$ | $WP_{11}$ | $WP_{19}$ | $WP_5$ | $WP_9$ | $WP_{13}$ | $WP_8$ | $WP_D$ | $\rightarrow$ | $WP_S$ | $WP_{11}$ | $WP_6$ | $WP_{19}$ | $WP_5$ | $WP_9$ | $WP_{13}$ | $WP_8$ | $WP_D$ |
|---|---|---|---|---|---|---|---|---|---|---|---|---|---|---|---|---|---|---|
| $WP_S$ | $WP_6$ | $WP_{11}$ | $WP_{19}$ | $WP_5$ | $WP_9$ | $WP_{13}$ | $WP_8$ | $WP_D$ | $\rightarrow$ | $WP_S$ | $WP_{13}$ | $WP_{11}$ | $WP_{19}$ | $WP_5$ | $WP_9$ | $WP_6$ | $WP_8$ | $WP_D$ |
| $WP_S$ | $WP_6$ | $WP_{11}$ | $WP_{19}$ | $WP_5$ | $WP_9$ | $WP_{13}$ | $WP_8$ | $WP_D$ | $\rightarrow$ | $WP_S$ | $WP_{13}$ | $WP_9$ | $WP_5$ | $WP_{19}$ | $WP_6$ | $WP_{13}$ | $WP_8$ | $WP_D$ |

**Fig.9.** Respectively the insertion, swap, and inversion mutations

### 4.5 Termination Criteria

The termination of the GA process is defined by completion of the maximum number of iterations, appearance of no change in population fitness after several iterations, and approaching to a stall generation.

### 4.6 Route Optimality Evaluation

The most important step in finding an optimum route by GA is forming an efficient cost function, so that the algorithm tends to compute best fitted solution with minimum cost value. The problem involves multiple objectives that should be satisfied during the optimization process. One approach in solving multi-objective problems is using multi-objective optimization algorithms. Another alternative is to transform a multi-objective problem into a constrained single-objective problem. The cost function for the route planner is defined as particular combination of weighted factors that are required to be maximized or minimized (given in section 6).

## 5 Overview of PSO and its Process on Path Planning

The PSO is one of the fastest optimization methods for solving variety of the complex problems and widely used in past decades. The argument for using PSO in path planning problem is strong enough due to its superior capability in scaling well with complex and multi-objective problems. The process of PSO is initialized with a population of particles. Each particle involves a position and velocity in the search space that get updated iteratively using equation (13). Each particle has memory to preserve the previous state values of best position $P^{P\text{-}best}$, as and the global best position $P^{G\text{-}best}$. The current state value of the particle is compared to the $P^{P\text{-}best}$ and $P^{G\text{-}best}$ in each iteration. More detail about the algorithm can be found in related references (Kennedy and Eberhart 1995).

**PSO Path Planning**

Initialize each particle by random velocity and position in following steps:
- Assign B-spline control points $\vartheta_i$ as particle position $\chi_i$
- Initialize each particle with random velocity $v_i$ in range of predefined bounds $\beta'_\vartheta = [U'_\vartheta, L'_\vartheta]$
- Choose appropriate parameters for the population size $nPop$
- Set the number of control-points ($n$) that used to generate the B-Spline path
- Set the maximum number of generations(iteration $Path_{Iter}$)
- Initialize the swarm population
- Initialize each particles current best position $P_i^{P\text{-}best}(1)$, at first iteration $t=0$
- Set the $P_i^{P\text{-}best}(1)$ as the best fitted particle $P_i^{G\text{-}best}(1)$ so far, at iteration $t$

BEGIN
Step-1: evaluate the cost function for each candidate particle (path) according to given cost function
Step-2: Updated the particles personal best $P_i^{P\text{-}best}$ and swarm's global best position $P_i^{G\text{-}best}$ at iteration $t$
for $i=1$ to $nPop$
$$P_i^{P\text{-}best}(t) = \begin{cases} P_i^{P\text{-}best}(t-1), & \text{if } \{Cost_\varphi(P_i(t)) \geq Cost_\varphi(P_i^{P\text{-}best}(t-1))\} \\ P_i(t), & \text{if } \{Cost_\varphi(P_i(t)) < Cost_\varphi(P_i^{P\text{-}best}(t-1))\} \end{cases}$$
$$P^{G\text{-}best}(t) = \underset{1 \leq i \leq P\text{-}Best}{\arg\min} Cost_\varphi(P_i^{P\text{-}best}(t))$$
end
Step-3: Update the state of the particle in the swarm
for $j=1$ to $i$
$\upsilon_{ij}(t) = \omega\upsilon_{ij}(t-1) + c_1 r_1 [P_{ij}^{P\text{-}best}(t-1) - \chi_{ij}(t-1)] + c_2 r_2 [P_{ij}^{G\text{-}best}(t-1) - \chi_{ij}(t-1)]$
$\chi_{ij}(t) = \chi_{ij}(t-1) + \upsilon_{ij}(t)$
end
Step-4: Repeat steps (1)–(3) for $t$ iterations or until the algorithm meets the stop criterion.
Step-5: Output $P^{G\text{-}best}$ as the optimal fitness value and its correlated path as the optimal solution
END

**Fig .10.** PSO optimal path planning pseudo code

$$\begin{aligned}
\upsilon_i(t) &= \omega\upsilon_i(t-1) + c_1 r_1 \left[P_i^{P\text{-}best}(t-1) - \chi_i(t-1)\right] + c_2 r_2 \left[P_i^{G\text{-}best}(t-1) - \chi_i(t-1)\right] \\
\chi_i(t) &= \chi_i(t-1) + \upsilon_i(t)
\end{aligned} \quad (13)$$

where $c_1$ and $c_2$ are acceleration coefficients, $\chi_i$ and $\upsilon_i$ are particle position and velocity at iteration $t$. $P_i^{P\text{-}best}$ is the personal best position and $P_i^{G\text{-}best}$ is the global best position. $r_1$ and $r_2$ are two independent random numbers in [0,1]. $\omega$ exposes the inertia weight and balances the PSO algorithm between the local and global search. Each particle in the swarm assigned by a potential path. The position and velocity parameters of the particles correspond to the coordinates of the B-spline control-points $\vartheta_i$ that utilises in path generation. The path planning is an optimization problem that aims to minimize the travel distance/time and avoid colliding obstacle(s). As the PSO algorithm iterates, every particle is attracted towards its respective local attractor based on the outcome of the particle's individual and swarm search results. The fitness of each generated path (particle) gets evaluated according to the fitness/cost functions discussed in section 6. All control points $\vartheta = \{\vartheta_1, \vartheta_2, ..., \vartheta_i, ..., \vartheta_n\}$ should be located in respective search region constraint to predefined bounds of $\beta^i{}_\vartheta = [U^i{}_\vartheta, L^i{}_\vartheta]$. If $\vartheta^i{:}(x^i, y^i, z^i)$ represent one control point in Cartesian coordinates in $t^{th}$ path iteration, $L^i{}_\vartheta$ is the lower bound; and $U^i{}_\vartheta$ is the upper bound of all control points at ($x$-$y$-$z$) coordinates given by (14):

$$\begin{aligned}
L_{\vartheta(x)} &= [x_0, x_1, x_2, ..., x_{i-1}, ..., x_{n-1}], & U_{\vartheta(x)} &= [x_1, x_2, ..., x_i, ..., x_n], \\
L_{\vartheta(y)} &= [y_0, y_1, y_2, ..., y_{i-1}, ..., y_{n-1}], & U_{\vartheta(y)} &= [y_1, y_2, ..., y_i, ..., y_n], \\
L_{\vartheta(z)} &= [z_0, z_1, z_2, ..., z_{i-1}, ..., z_{n-1}], & U_{\vartheta(z)} &= [z_1, z_2, ..., z_i, ..., z_n],
\end{aligned} \quad (14)$$

With respect to given relations (14), each control point is generated from (15):

$$\begin{aligned}
x^i(t) &= L^i_{\vartheta(x)} + Rand^x_i (U^i_{\vartheta(x)} - L^i_{\vartheta(x)}) \\
y^i(t) &= L^i_{\vartheta(y)} + Rand^y_i (U^i_{\vartheta(y)} - L^i_{\vartheta(y)}) \\
z^i(t) &= L^i_{\vartheta(z)} + Rand^z_i (U^i_{\vartheta(z)} - L^i_{\vartheta(z)})
\end{aligned} \quad (15)$$

where ($x_0, y_0, z_0$) and ($x_n, y_n, z_n$) are the position of the *start* and *target* points in the LPP, respectively. The pseudo code of the PSO algorithm and its mechanism on path planning process is provided in *Fig.*10.

## 6 Architecture Evaluation

AUV starts its mission from start point and should serve sufficient number of tasks to reach the destination on-time. Given a candidate route in a sequence of waypoints along with environmental information, the LPP module provides a trajectory to safely guide the vehicle through the waypoints. The resultant local path should be time optimum, safe and flyable (feasible). It shouldn't cross the forbidden area covered by obstacles $\Theta$ (defined using eq 2-5). If the $\vartheta = \{\vartheta_1, \vartheta_2, ..., \vartheta_i, ..., \vartheta_n\}$ is the sequence of control points along each arbitrary local path from set of $\wp = \{\wp_1, \wp_2, ...\}$, the path $\wp_i$ gets evaluated by a cost function $Cost_\wp$ defined based on travel time $t_i \approx T_{path\text{-}flight}$ required to pass the path segments. The route cost has direct relation to the passing distance among each pair of selected waypoints. Hence, the path cost $Cost_\wp$ for any optimum local path get used in the context of the GRP. The model is seeking an optimal solution in the sense of the best combination of path, route, and task cost. The route function $Cost_{Route}$ gets penalty when the $T_{Route}$ for a particular route $R_i$ exceeds the $T_{Available}$. Thus, the provided route and path is evaluated as follows.

$$\forall \wp, \quad \wp \approx \sum_1^{|\wp|} \vartheta_{j+1} - \vartheta_j, \quad \wp^j_{x,y,z} = \sum_{x_s, y_s, z_s}^{|\wp|} \sqrt{(\vartheta_{x(j+1)} - \vartheta_{x(j)})^2 + (\vartheta_{y(j+1)} - \vartheta_{y(j)})^2 + (\vartheta_{z(j+1)} - \vartheta_{z(j)})^2} \quad (16)$$

$$T_{path\text{-}flight} = \sum_1^n t_i = \sum_1^{|\wp|} \Gamma_{3D}\{\vartheta^i_{t_i}\} = \sum_1^{|\wp|} \frac{|\vartheta^{t_i}_{i+1} - \vartheta^{t_i}_i|}{|V_{AUV}|} \quad (17)$$

$$\begin{aligned}
&Cost_\wp \approx \min(T_{path\text{-}flight}) \\
&s.t. \\
&\forall j \in \{0, ..., |\wp|\} \Rightarrow \vartheta^{t_i}_j \notin \Theta(t_j) \bigcup \Gamma_{3D} \quad and \quad j \notin \bigcup_{N\Theta} \Theta(\Theta_p, \Theta_r, \Theta_{Ur})
\end{aligned} \quad (18)$$

$$\begin{aligned}
&Cost_{Route} \propto |T_{Route} - T_{Available}| \\
&Cost_{Route} \propto \left| \sum_{\substack{i=0 \\ j \neq i}}^n lq_{ij} \left(\frac{Cost_{\wp ij}}{V_{AUV}} + \delta_{Tij}\right) - T_{Available} \right|, \quad l \in \{0,1\} \\
&s.t. \\
&\forall R_i \Rightarrow \max(T_{Route}) < T_{Available}
\end{aligned} \quad (19)$$

$$Cost_{Task} \propto \sum_{\substack{i=0 \\ j \neq i}}^n lq_{ij} \left(\frac{\xi_{Route}}{\rho_{Route}}\right), \quad l \in \{0,1\} \quad (20)$$

After visiting each waypoint, the re-planning criterion is investigated. A computation cost encountered any time that re-planning is required. Thus, the total cost for the model defined as:

$$Cost_{Total} = \varphi_1 Cost_{Task} + \varphi_2 Cost_{Route}(Cost_\wp) + \sum_1^r T_{compute} \quad (21)$$

where $T_{compute}$ is the time required for checking the re-planning criteria and computing the new optimum route, and $r$ is the number of repeating the re-planning procedure. $N_\Theta$ is number of obstacles. $\Theta_p$, $\Theta_r$, and $\Theta_{Ur}$ are obstacle position, radius and uncertainty, respectively. $\varphi_1$ and $\varphi_2$ are two positive numbers that determine amount of participation of $Cost_{Task}$ and $Cost_{Route}$ on calculation of total cost $Cost_{Total}$. Giving the appropriate value for coefficients of engaged factors in the cost function has a significant effect on performance of the model.

## 7 Results and Discussion

The main purpose of this research is evaluating the performance of entire architecture in terms of increasing mission productivity (task assignment and time management), while guaranteeing vehicles safety during the mission. To verify the efficiency of the proposed architecture, the performance of each module is investigated individually from top to bottom layer and explained in following subsections.

## 8 Simulation Results and Discussion

The main purpose of this research is evaluating the performance of entire architecture in terms of increasing mission productivity (task assignment and time management), while guaranteeing vehicles safety during the mission. To verify the efficiency of the proposed architecture, the performance of each module is investigated individually from top to bottom layer and explained in following subsections.

### 8.1 Simulation Results for Methods Used in GRP Module

At the top level of the architecture, a configurable GRP module is developed in order to find the most productive optimum global route between start and destination points. Two different algorithms are adapted and tested by module to evaluate the optimality of the global route. Several different criteria are embedded to keep the generated routes concentrated to the feasible solution space, which comprehensively reduces the memory usage and time complexity of the searching process. The global route gives a general overview of the area that AUV should fly through by cutting off the operation area to smaller beneficent zone for vehicle's deployment. The GRP operates based on offline map information and does not deal with dynamic changes of terrain. Assumptions for GRP module are given below.

i) In this study, it is assumed that vehicle is moving in a 3D environment covered by multiple fixed waypoints that one of them is the start point which vehicle starts its mission from that and one of them is destination point (dock for example) that vehicle should reach to that point within mission available time. This information represented in a graph form terrain.
ii) Tasks assigned to edges of the graph in advance. Each task involves three parameters of priority, risk percentage and required completion time. AUV is moving with static velocity and is requested to serve maximum number of tasks in mission time.

To evaluate efficiency the GRP module for a single vehicle routing problem, its performance in task allocation, time management, productivity of the mission, real-time performance, and other factors are tested using two different evolutionary strategies of GA and Imperialist Competitive Algorithm (ICA), which both are popular meta-heuristic optimization methods in solving NP hard problems. More detail about ICA optimization algorithm can be found in (Movahed et al. 2011; Soltanpoor et al. 2013). A number of performance metrics have been investigated to evaluate the quality/optimality of the proposed solutions in different network topologies. One of these metrics is the reliability percentage of the route representing chance of mission success, which is combination of route validity to time restriction and feasibility criteria. Other metrics involve the number of completed tasks, total obtained weight, total cost, and the time optimality of the generated route with respect to complexity of the graph. These metrics altogether perform single vehicles mission productivity in a specific time interval. The ICA and GA configured with the same initial conditions of 150 iterations and 100 populations. The performance of both algorithms is tested on two graphs with the same complexities, one with 50 nodes and another one with 100 nodes, presented in Table 1 and *Fig*.11.

Table 1. Statistical analyzing of route evaluation for two different graph complexity for both ICA and GA

| *Performance Metrics* | | *Topology 1* | | *Topology 2* | |
|---|---|---|---|---|---|
| ***Number of Nodes*** | | 50 Nodes | | 100 Nodes | |
| ***Number of Edges*** | | 1197 | | 4886 | |
| ***Algorithm*** | | **ICA** | **GA** | **ICA** | **GA** |
| ***CPU Time(sec)*** | | 18.4 | 9.5 | 20.2 | 17.53 |
| ***Best Cost*** | | 0.056 | 0.034 | 0.047 | 0.029 |
| ***Available Time(sec)*** | | 25200 (7h) | 25200 (7h) | 25200 (7h) | 25200 (7h) |
| ***Route Time(sec)*** | | 22218 | 23981 | 25212 | 23875 |
| ***Total Distance*** | | 55329 | 61812 | 669857 | 63417 |
| ***Total Weight*** | | 45 | 54 | 47 | 57 |
| ***N-Tasks*** | | 16 | 19 | 18 | 23 |
| ***Reliability*** | *Violation* | 0.00 | 0.00 | 0.0043 | 0.00 |
| | *Feasibility* | Yes | Yes | Slightly late | Yes |

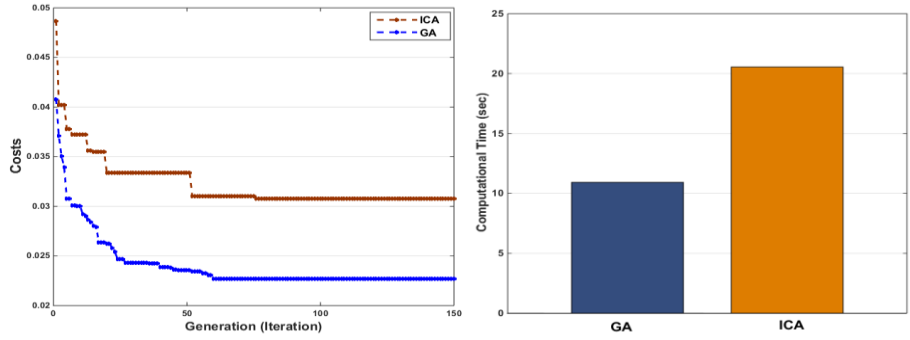
**Fig.11.** (a) GA and ICA cost variations in 150 iterations, (b) GA and ICA total computational time in 150 iterations

From simulation results in Table.1, it is noted that in all cases route travelling time obtained by GA is smaller than total available time and violation value for all solutions is equal to zero that confirms feasibility of the produced route, which means GA acts according to defined constraints. It is clear from Table.1 and *Fig*.11 that GA acts more efficiently in terms of minimizing cost value and computation time comparing to ICA. The provided results also confirms the superior performance of the GA based route planner in terms of increasing mission productivity by maximizing total obtained weight and number of covered tasks by taking maximum use of the available time (as $T_{Route}$ considerably approaches the $T_{Available}$). Indeed it is evident from Table.1 and *Fig*.11 that the performance of both algorithms is relatively independent of both size and complexity of the graph, as this is a challenging problem for other deterministic algorithms. Hence, the evolutionary algorithms are suitable to produce optimal solutions quickly for real-time applications.

To evaluate the stability and reliability of the employed algorithms in terms of total route time, CPU time, distance, and total obtained weight, 100 execution runs are performed in a Monte Carlo simulation, presented by *Fig*.12.

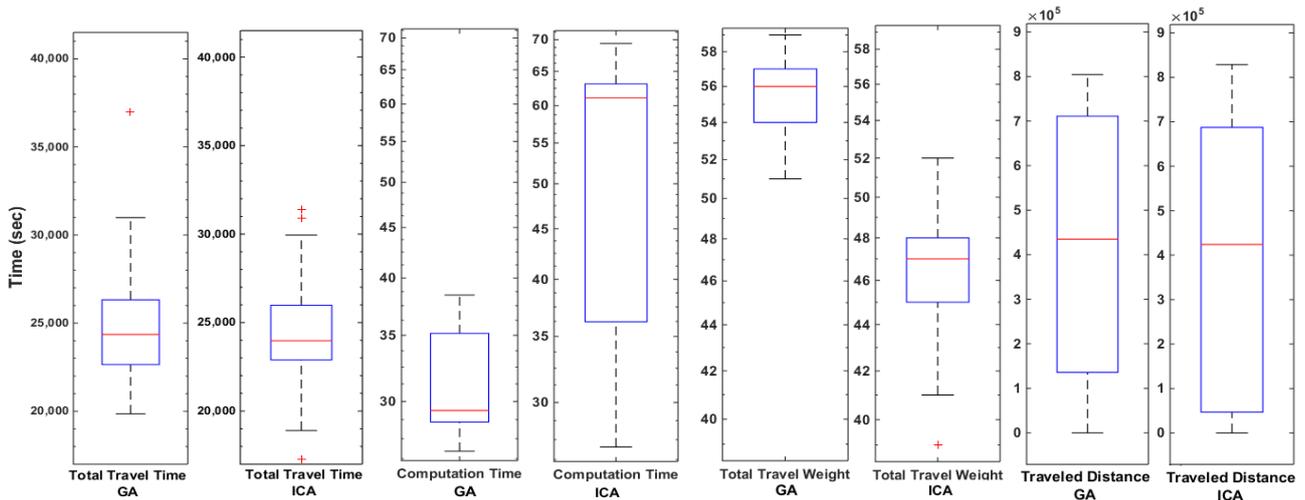
**Fig.12.** Comparison of stability of GA and ICA in terms of satisfying given performance metrics based on Monte Carlo simulation

The number of graph nodes is fixed on 20 waypoints for all Monte Carlo runs, but the topology of the graph was changed randomly based on a Gaussian distribution on the problem search space. The time threshold ($T_{Available}$) also fixed on $2.52 \times 10^4 (sec)$. *Fig*.12 compares the functionality of GA and ICA in dealing with problem's space deformation and quantitative measurement of four significant mission's metrics of travel time, CPU time, total weight, and total traveled distance. As indicated in *Fig*.12, GA has superior performance and shows more consistency in its distribution comparing to the generated solutions by ICA algorithm. The GA reveals robust behavior to the variations and meet the specified constraint.

8.2  Simulation Results for PSO-based planner Used in LPP Module

The path planning is an optimization problem in which the main goal is to minimize the travel distance and time $T_{path-flight}$, and avoiding colliding obstacle(s). The following assumptions are considered in generation local optimum path.

a) The ocean current velocity is assumed to be constant. As the path planner is applied in a small scale area, the water current has effect on both floating and moving obstacles, where moving obstacles have self-motivated velocity additional to current velocity. The floating obstacles considered with a growing uncertainty rate $\Theta_{Ur}$ proportional to current velocity ($U_R^C(t) \sim |V_C|$).

b) The AUV considered to have constant thrust power, and therefore, the battery usage for a path is a constant multiple of the distance travelled. Therefore, it is assumed the AUV travelling with constant velocity of $V_{AUV}$.

These assumptions play important role in efficient path planning and copping with terrain dynamic changes. In path planning simulation the obstacles are generated randomly from different categories and configured individually based on given relations in section 3. Encountering different type of obstacles, this research investigates four different scenario in terms of the dynamicity of the environment.

*Scenario-1*: The AUV starts its deployment in a pure static operating filed covered by random combination of the known static and uncertain static obstacles, in which obstacles are under the exposure of varying levels of position uncertainty propagating from the center of the obstacle. The vehicle is required to pass the shortest collision free distance to reach to the specified target waypoint.

*Scenario-2:* Making the AUV's deployment more challenging, in the second scenario, the robustness of the method is tested in a dynamic environment with moving obstacles, in which obstacle position changes to a random direction by uncertainty rate proportional to time, where the number of obstacles increases by time.

*Scenario-3:* In the third scenario, the mission becomes more complicated by encountering the current force on moving obstacles with uncertain position, in which the obstacle has self-motivated velocity to a random direction and affected by current force that presented with a growing uncertainty proportional to the current velocity $U_R^C \sim |V_C|$ radiating out from centre of the object.

*Scenario-4:* The last case, an irregularly shaped terrains including all static, floating, and moving obstacles encountered in computing optimum trajectory.

All four scenarios simulated for varying number of 3 to 6 obstacles in corresponding operation window. The purpose of this simulation is evaluating the ability of the proposed method in balancing between searching unexplored environment and safely swimming toward the target waypoint. For this purpose, a distinctive number of runs are performed to analyze the performance of the method in satisfying the problem constrains for all mentioned scenarios. The PSO optimization configuration set by 150 particles (candidate paths) and 100 iterations. The expansion-contraction coefficients also set on *2.0* to *2.5*. The maximum number of control points for each B-spline is fixed on 8. The vehicles water-referenced velocity considered 3 m/s. Figure.13 represent the produced optimum trajectory in first scenario encountering 3 to 6 obstacles. The gradual increment of collision boundary is presented by *circle(s)* around the obstacles, in which the uncertainty propagation is assumed to be linear with iteration/time. The performance of the algorithm in minimizing the cost and eliminating the violation is for all scenarios represented by *Fig*.13(b-c) to *Fig*.15(b-c). The purpose of increasing the number of obstacles is to evaluate sustainability of the path planning performance to complexity of the terrain.

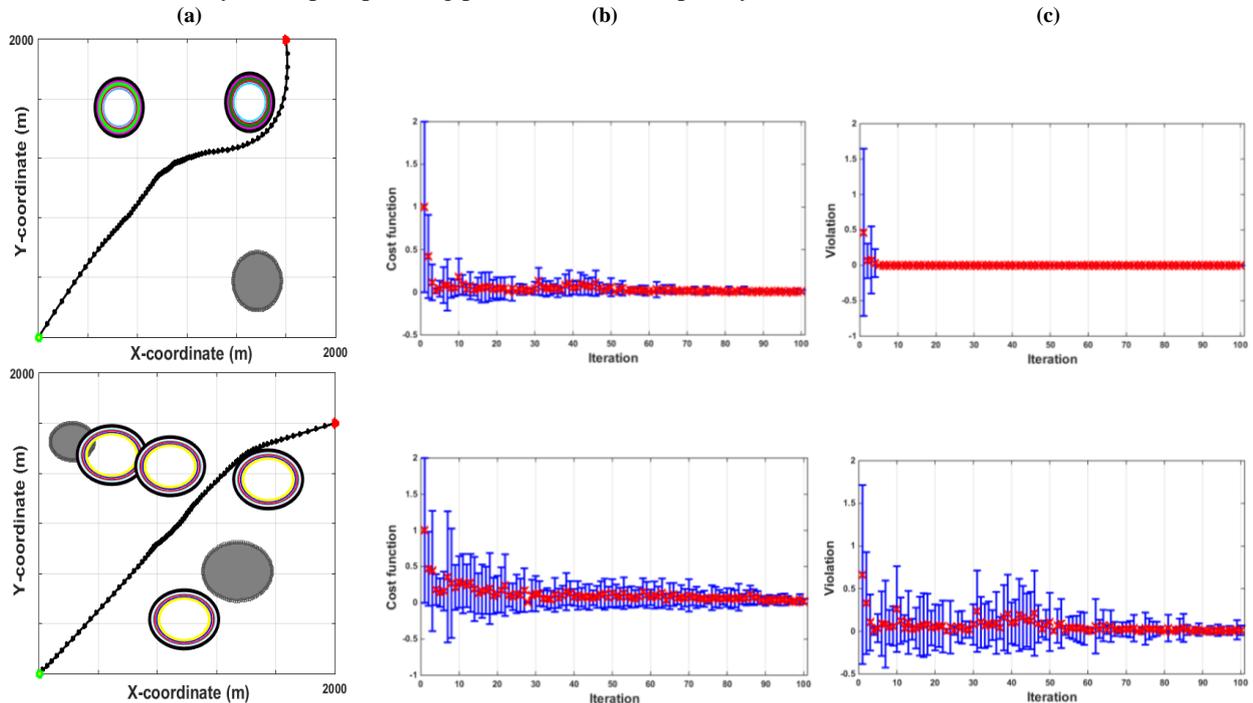

**Fig.13.** (a) 2-D representation of generated optimum 3-D trajectory in *scenario-1*, including random combination **3** to **6** static known and static uncertain obstacles. (b) Cost variation of path population in each iteration. (c) Violation variation of path population in each iteration as collision penalty.

The performance of the algorithm investigated for the second scenario and presented by Fig.14, while number of obstacles is increased to 6. The obstacles movement also occurs in a specified rate of uncertainty proportional to time in a random direction.

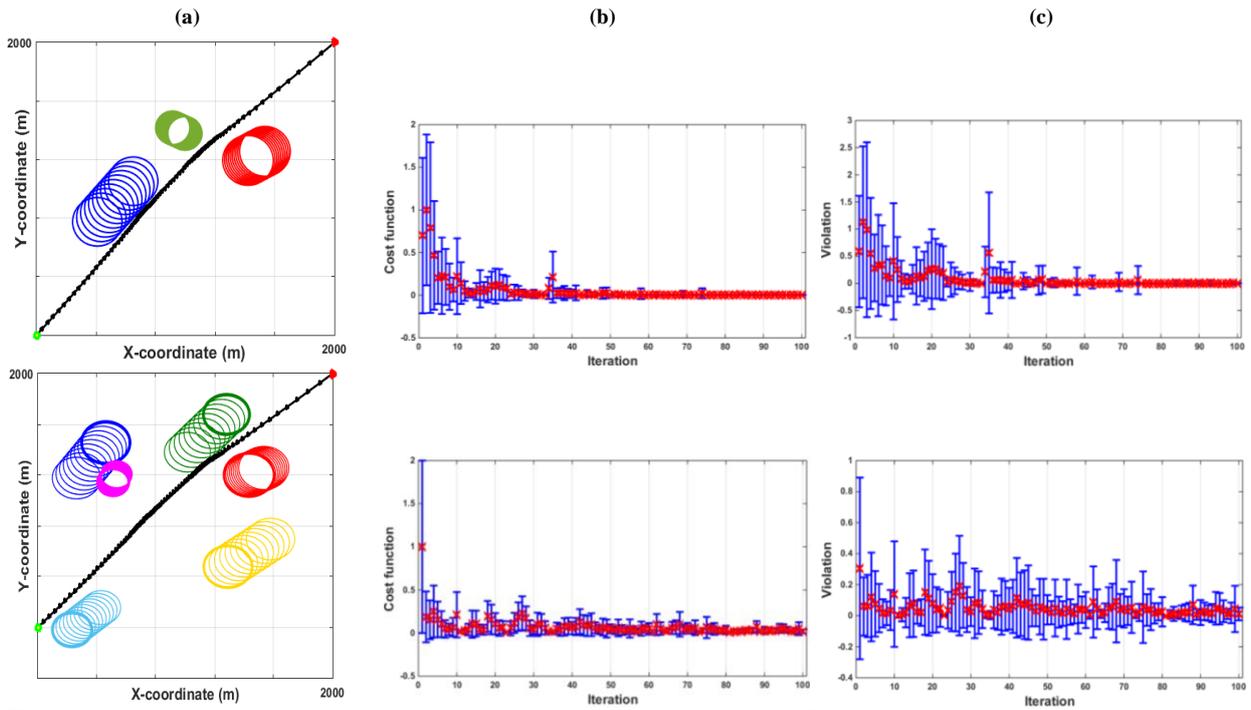

**Fig.14.** (a) 2-D representation of generated optimum 3-D trajectory in *scenario-2*, including **3** to **6** moving obstacles. (b) Cost variation of path population in each iteration. (c) Violation variation of path population in each iteration as collision penalty.

Figure 15 represents the produced optimum trajectory for 3 to 6 obstacles in the third scenario, respectively. The uncertainty around the obstacles propagates from the centre of the object in all directions with a growth rate proportional to current velocity. Additionally, the obstacles move with a self-motivated velocity in a random direction.

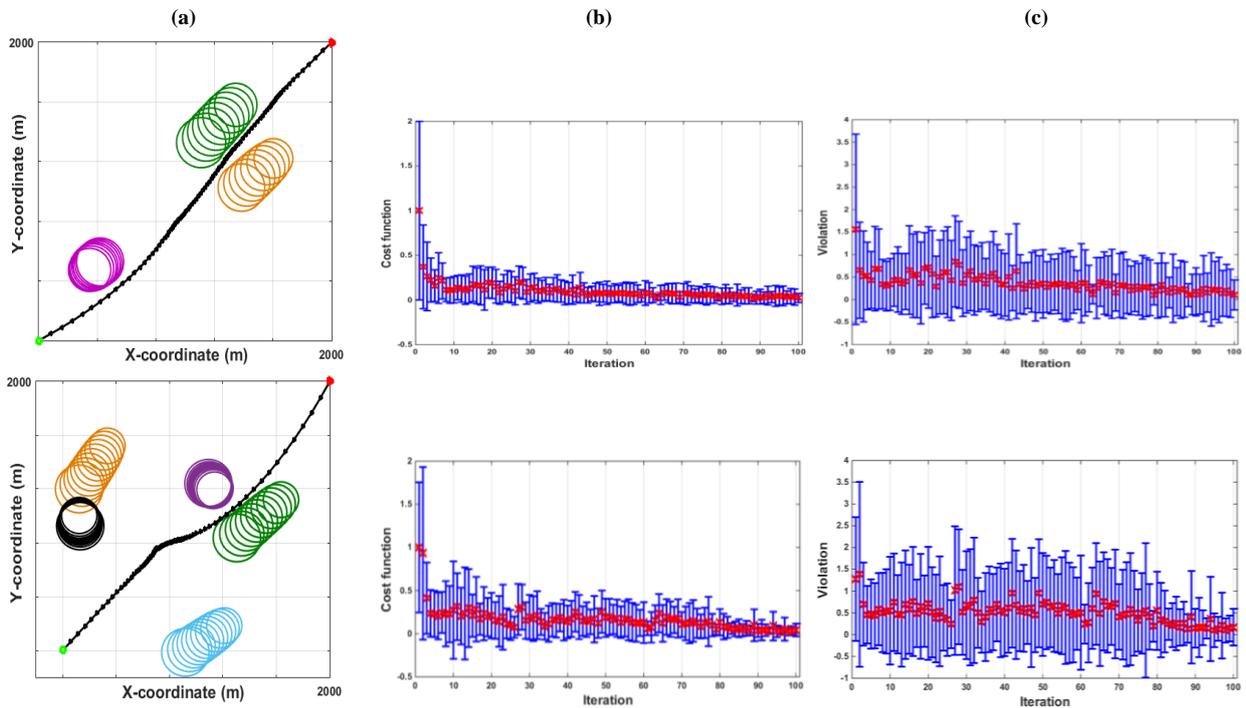

**Fig.15.** (a) 2-D representation of generated optimum 3-D trajectory in *scenario-3*, including **3** to **6** obstacles. (b) Cost variation of path population in each iteration. (c) Violation variation of path population in each iteration as collision penalty.

Referring to *Fig*.13(b), 14(b) and 15(b), it is evident that the path population converges to the minimum cost by passing iterations. The cost variation range decreases in each iteration which means algorithm accurately converges the solution space to the optimum solution. The red crosses in the middle of the bar charts represent the mean cost of path population in each iteration. Tracking the variation of the mean cost and mean violation in *Fig*.13(b,c), 14(b,c) and 15(b,c) declares that algorithm accurately pushes the solutions to approach the optimum solution with minimum cost and efficiently manages the trajectory to eliminate the collision penalty within 100 iterations.

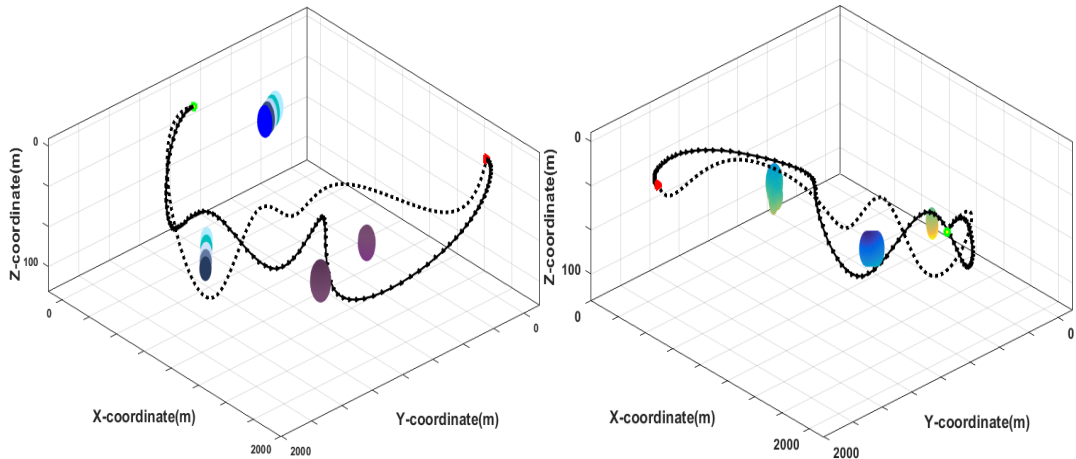

**Fig.16.** The generated trajectory in *scenario-4* with random composite of all four types of obstacles

The simulation result for last scenario is provided in *Fig*.16 in which the performance of the proposed method in generating collision free shortest trajectory is investigated for a random combination of all types of obstacles. The trajectory is plotted in 3-d format for clear graphical representation of its collision avoidance capability. The simulation results represented in *Fig*.13 to *Fig*.16 shows that the proposed path planning method accurately generates collision free time optimal trajectories and dynamically adapts to environmental changes encountering uncertain, static, floating, and moving obstacles. Increasing the number of obstacles, increases the problems complexity, however, it is derived from results that the performance of the algorithm is almost stable against increasing the complexity of the terrain and the algorithm tends to minimize the travel distance and time, which furnishes the expectation of the architecture at lower level of the autonomy. Any time that LPP is recalled from the main model, it dynamically computes optimum path based on observed change in the environment and new obtained information. The AUV travels through the listed waypoints in optimum global route with 3m/s water-referenced velocity, and passes waypoints one by one, in a way that a target waypoint for LPP, would be a new start position in next run. This process repeats until vehicle reaches to the final destination. Therefore, the initial and destination waypoints and operation field for LPP changes as vehicle passes through the waypoints in global route sequence. The next step is evaluation of the entire model in terms of appropriate decision making and providing efficient interaction and cooperation between the high and low level modules. Additional to addressed common performance indexes discussed above, two other factors are highlighted for the purpose of this research that are investigated along the evaluation of the architectures performance. The first critical factor for the LPP is the computational time. The second important factor considered for the purpose of this research is the existence of reasonable and close correlation between generated path time ($T_{path\text{-}flight}$) the expected time ($T_{Expected}$), which is investigated meantime the evaluation of the architecture along the checking process for requisition of re-planning.

### 8.3 Architecture Performance Evaluation

In this section, the simulation result of the proposed configurable architecture for AUV's mission management is presented. The main architecture aims to take the maximum use of the mission available time, to increase the mission productivity by optimum routing, and guarantee on-time termination of the mission; and concurrently ensuring the vehicles safety by copping dynamic unexpected challenges during deployment toward the final destination. Accurate synchronization of the inputs and outputs to the main model and concurrent cooperation of the engaged modules are the most important requirements in stability of the architecture toward the main objectives addressed above. To this end, the robustness of the model in enhancement of the vehicles autonomy is evaluated by testing 10 missions' through 10 individual experiments presented in *Fig*.17 to *Fig*.19.

The initial configuration of the operation network has been set on 50 waypoints and 1470 edges involving a fixed sequence of tasks with specified characteristics (priority, risk percentage, completion time) in 10 $km^2$(*x-y*), 1000 *m*(*z*) space. The waypoints location are randomized according to $\sim U(0,10000)$ for $WP^i_{x,y}$ and $\sim U(0,100)$ for $WP^i_z$. The mission available time for all experiments is fixed on $T_{Available}=10800$(*sec*)=3 (*hours*). The vehicle starts its mission at initial location $WP^1$ and ends its mission at $WP^{50}$. The operating field is modeled as a realistic underwater environment that randomly covered by different uncertain static, floating and moving obstacles, where the floating obstacles is affected by current force varied according to $|V_C|\sim N(0,0.3)$. For the purposes of this study, the optimization problem was performed on a desktop PC with an Intel i7 3.40 GHz quad-core processor in MATLAB® R2014a. The LPP as an inner component operates in context of the GRP module and output of each module concurrently feeds to another one. One mission progress has been provided in Table.2 (A-B) to clarify the process of the architecture in different stages of a specific mission toward carrying out the mentioned objectives.

Table.2. Process of the architecture in one mission scenario

| A. Global Route Planning (GRP) Module | | | | | | | | | |
|---|---|---|---|---|---|---|---|---|---|
| Call NO | Start | Dest | Task NO | Weight | Cost | CPU | $T_{Available}$ | $T_{Route}$ | Validity | Route Sequence |
| 1 | 1 | 50 | 8 | 22 | 0.048 | 17.3 | 10800 | 10262 | Yes | 1-39-7-16-48-33-40-38-50 |
| 2 | 7 | 50 | 10 | 34 | 0.030 | 23.9 | 7710.8 | 7805 | Yes | 7-41-14-12-36-48-22-15-47-40-50 |
| 3 | 12 | 50 | 7 | 38 | 0.024 | 21.9 | 5375.3 | 5211 | Yes | 12-4-39-44-30-28-11-50 |
| 4 | 4 | 50 | 5 | 44 | 0.040 | 19.8 | 4234.7 | 3998 | Yes | 4-41-12-44-11-50 |
| 5 | 41 | 50 | 5 | 36 | 0.029 | 18.4 | 3474.7 | 3468 | Yes | 41-46-3-44-17-50 |
| 6 | 3 | 50 | 4 | 38 | 0.078 | 20.7 | 2117.3 | 2008 | Yes | 3-29-30-42-50 |
| 7 | 30 | 50 | 2 | 35 | 0.477 | 21.8 | 1054.5 | 1051 | Yes | 30-42-50 |

| B. Local Path Planning (LPP) Module | | | | | | | | | | |
|---|---|---|---|---|---|---|---|---|---|---|
| Route ID | PP Call | Edges | Violation(Collision) | Cost | CPU | $T_{path\text{-}flight}$ | $T_{Expected}$ | $T_{Available}$ | Replan Flag | LPP Flag |
| Route-1 | 1 | 1-39 | 0.000000 | 0.2260 | 47.3 | 2333.1 | 2535.3 | 8466.7 | 0 | 1 |
|  | 2 | 39-7 | 0.000043 | 0.7010 | 39.8 | 755.8 | 666.6 | 7710.8 | 1 | 0 |
| Route-2 | 1 | 7-41 | 0.000000 | 0.1460 | 42.4 | 501.4 | 508.3 | 7209.4 | 0 | 1 |
|  | 2 | 41-14 | 0.000000 | 0.3170 | 40.0 | 1078 | 1179 | 6131.4 | 0 | 1 |
|  | 3 | 14-12 | 0.000000 | 0.2260 | 42.3 | 756.1 | 686.6 | 5375.3 | 1 | 0 |
| Route-3 | 1 | 12-4 | 0.000000 | 0.2790 | 40.9 | 1140.6 | 528.3 | 4234.7 | 1 | 0 |
| Route-4 | 1 | 4-41 | 0.000000 | 0.2020 | 37.4 | 760.02 | 696.8 | 3474.7 | 1 | 0 |
| Route-5 | 1 | 41-46 | 0.000000 | 0.2017 | 40.6 | 674.5 | 820.6 | 2800.2 | 0 | 1 |
|  | 2 | 46-3 | 0.000000 | 0.2037 | 44.1 | 682.8 | 647.8 | 2117.3 | 1 | 0 |
| Route-6 | 1 | 3-29 | 0.000000 | 0.1600 | 39.8 | 567.2 | 857.6 | 1550.1 | 0 | 1 |
|  | 2 | 29-30 | 0.000000 | 0.1460 | 43.4 | 495.5 | 334.8 | 1054.5 | 1 | 0 |
| Route-7 | 1 | 30-42 | 0.000000 | 0.1420 | 39.7 | 479.04 | 482.1 | 575.4 | 0 | 0 |
|  | 2 | 42-50 | 0.000000 | 0.1370 | 40.1 | 563.7 | 569.3 | **11.7** | 0 | 0 |

The mission starts with calling the GRP for the first time. The GRP produces a valid optimum route to take maximum use of available time (valid route $T_{Route} \leq T_{Available}$). Referring Table.2(A), the initial optimum route covers number of 8 tasks with total weight of *22* and cost of *0.048* with estimated completion time of $T_{Route}$=10262(*sec*). In the second step, the LPP is recalled to generate optimum collision free trajectory through the listed sequence of waypoint included in the initial route. Referring to Table.2(B), the LPP module got the first pair of waypoints (1-39) and generated optimum trajectory between location of $WP^1$ to location of $WP^{39}$ with total cost of the 0.2260, and travel time of $T_{path\text{-}flight}$=2333.1 which is smaller than $T_{Expected}$=2535.3. The $T_{Expected}$ for the LPP is calculated based on estimated travel time for the generated route $T_{Route}$. In cases that $T_{path\text{-}flight}$ is smaller than $T_{Expected}$ the re-planning flag is zero which means the initial optimum route is still valid and optimum, so the vehicle is allowed the follow the next pair of waypoints included in initial optimum route. After each run of the LPP, the $T_{path\text{-}flight}$ is reduced from the total available time $T_{Available}$. The second pair of waypoints (39-7) is shifted to the LPP and the same process is repeated. However, if $T_{path\text{-}flight}$ exceeds the $T_{Expected}$ re-planning flag gets one, which means some of the available time is wasted in passing the distance between $WP^{39}$ and $WP^7$ due to copping collision avoidance. In such a case also the $T_{Available}$ gets updated and visited edges (1-39, and 39-7) get eliminated from the graph. Afterward, instead of LPP, the GRP is recalled to generate new optimum route from the current waypoint $WP^7$ to the predefined destination $WP^{50}$ according to updated operation network and $T_{Available}$. In experimental results presented in Table.2, the GRP is recalled for 7 times and the LPP called for 13 times within 7 optimum routes. This synchronization among the modules continues until vehicle reaches to the destination (success) or $T_{Available}$ gets a minus value (failure: vehicle runs out of battery). The final route passed by the vehicle in this mission through the 7 route re-planning and 13 path planning is the sequence {1-39-7-41-14-12-4-41-46-3-29-30-42-50} with total cost of 0.038, total weight of 38, and total time of 10788.3.

The most appropriate outcome for a mission is completion of the mission with the minimum remained time, which means maximizing the use of mission available time. Referring Table.2(B), the remaining time is *11.7* out of the whole mission available time of $T_{Available}$=10800(*sec*)=3(*h*), which is considerably approached to zero. Therefore the architectures performance can be represented by mission time (or remained time) along with productivity of the mission by completing the maximum number of highest priority tasks with minimum risk percentage.

Considering the fact that reaching to the destination, as a big concern for vehicles safety, is more important than maximizing the vehicles productivity, a big penalty value is assigned to GRP to strictly prevent generating routes with $T_{Route}$ bigger than $T_{Available}$. To measure the performance of the proposed dynamic architecture in a quantitative manner, the robustness and stability of the model in enhancement of the vehicles autonomy in terms of mission time management and vehicles safety is evaluated through testing 10 individual missions with the same initial condition that closely matches actual underwater mission scenarios that presented in *Fig*.17 to *Fig*.19.

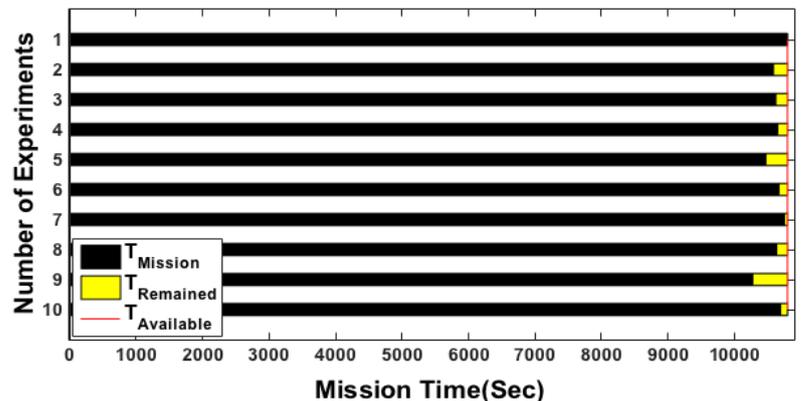

**Fig.17.** Architecture performance in maximizing mission's productivity by maximizing the mission time constraint to available time threshold and its computational stability

The stability of the architecture in time management is the most critical factor representing robustness of the method. It is derived from simulation results in *Fig*.17 that the proposed architecture is capable of taking maximum use of mission available time as apparently the mission time in all experiments approach the $T_{Available}$ and meet the above constraints denoted by the upper bound of 10800 *sec=3 hours* (is shown by red line). Respectively, the value of the remaining time that has a linear relation to $T_{Available}$, should be minimized but it should not be equal to zero which is accurately satisfied considering variations of remaining time for 10 experiments in *Fig*.17. In other words, minimizing the remaining time maximizes the mission productivity. To establish appropriate cooperation between the high and low level modules (GRP and LLP), the correlation between path time ($T_{path-flight}$) the expected time ($T_{Expected}$) is another important performance index investigated and presented in *Fig.18*.

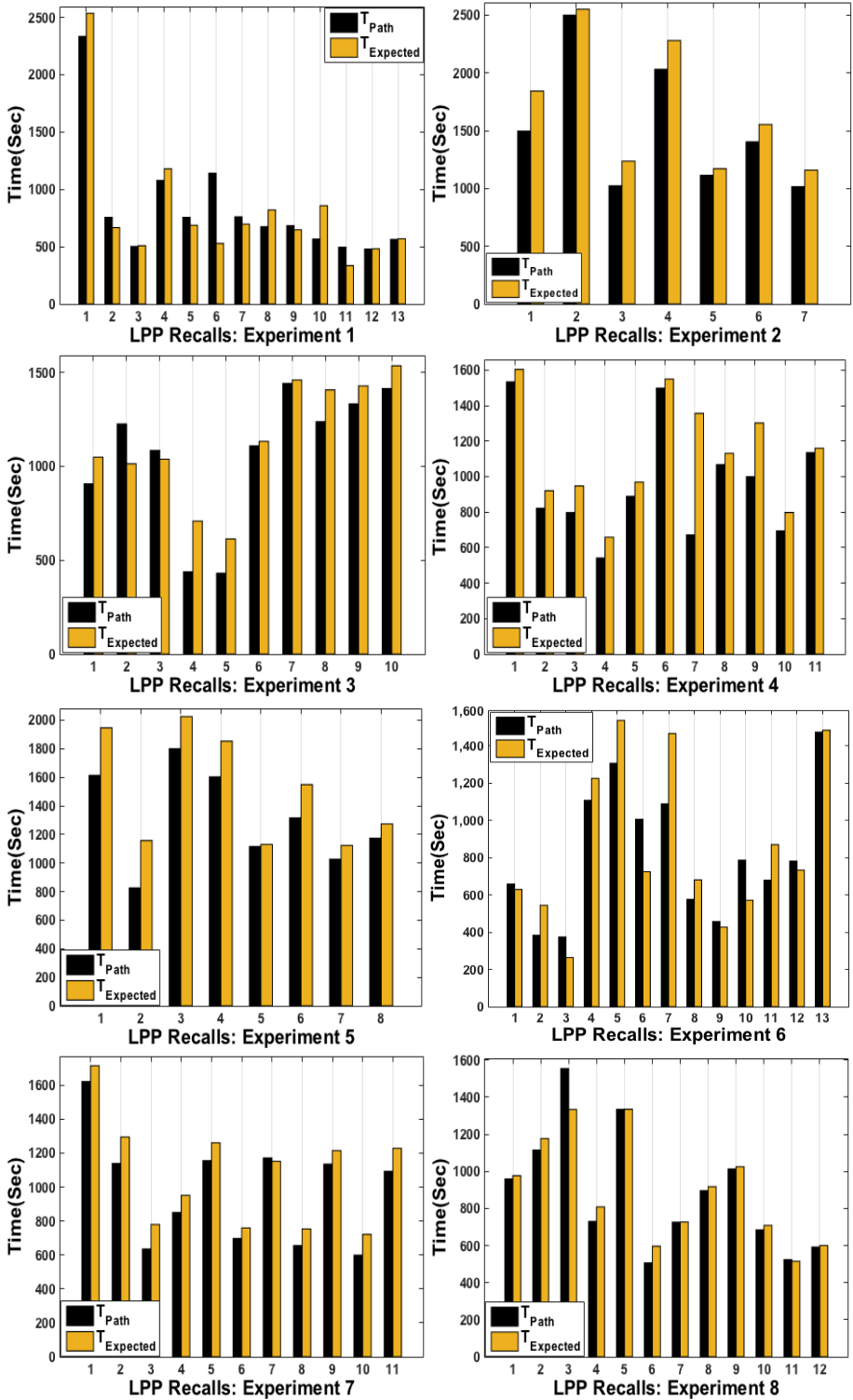

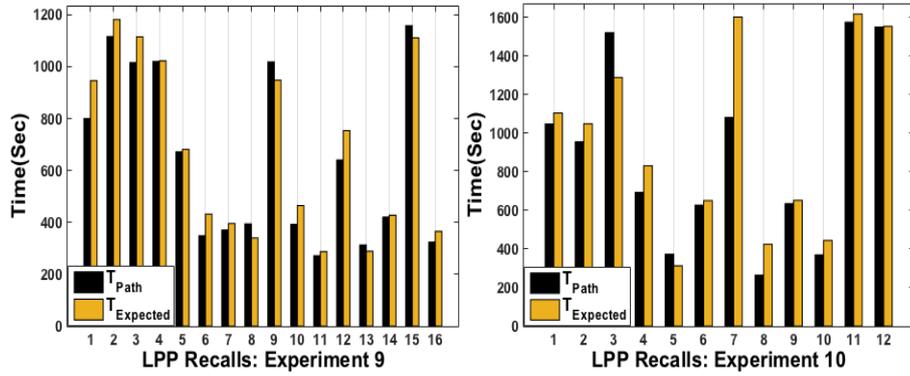
**Fig.18.** Stability of architecture in managing correlation of $T_{path\text{-}flight}$ and $T_{Expected}$ in multiple recall of LPP in 10 experiments

Figure 18 presents relation between value of $T_{path\text{-}flight}$ and $T_{Expected}$ in multiple recall of LPP in 10 different experiments. Existence of a reasonable difference between $T_{path\text{-}flight}$ and $T_{Expected}$ values in each LPP operation is critical to total performance of the architecture. In other words, there shouldn't be a big difference between these two parameters to prevent interruption in cohesion of the whole system. As discussed earlier, route replanning is required when the $T_{path\text{-}flight}$ exceeds $T_{Expected}$; hence, according to *Fig.*18, the Synchro-module is recalled for six times in mission 1, not recalled in mission 2, two times in mission 3, not recalled in mission 4 and 5, six times in mission 6, ones in mission 7, three times in mission 8, four times in mission 9, and two times in mission 10 in order to apply mission updates and carry out the replanning process. Another critical factor is the computational time for both LPP and GRP operations. The LPP must operate concurrently and synchronous to the GRP, thus a large computational time causes the LPP drop behind the operation of the GRP, which flaws the routine flow and cohesion of the whole system. Figure 19 presents the computational time for both LPP and GRP operations in multiple recalls through the 10 mission executions in boxplot format.

It is noteworthy to mention from analyze of results in *Fig.*19 that the proposed methodology takes a very short computational (CPU) time for all experiments that makes it highly suitable for real-time application. Besides, referring to Fig.19 it can be inferred that the variation of computational time is settled in a narrow bound (approximately in range of second for all experiments) for both GRP and LPP modules that confirm applicability the model for real-time implementation. Considering the indexes of the total mission time, remaining time, and variations of $T_{path\text{-}flight}$ and $T_{Expected}$, the results obtained from analyze of 10 different missions are quantitatively very similar that proves the inherent stability of the model. More importantly, the violation percentage in both GRP and LPP simulations presented in Table.2, reveal that both planners' are robust to the variations of the operation network parameters and environmental conditions.

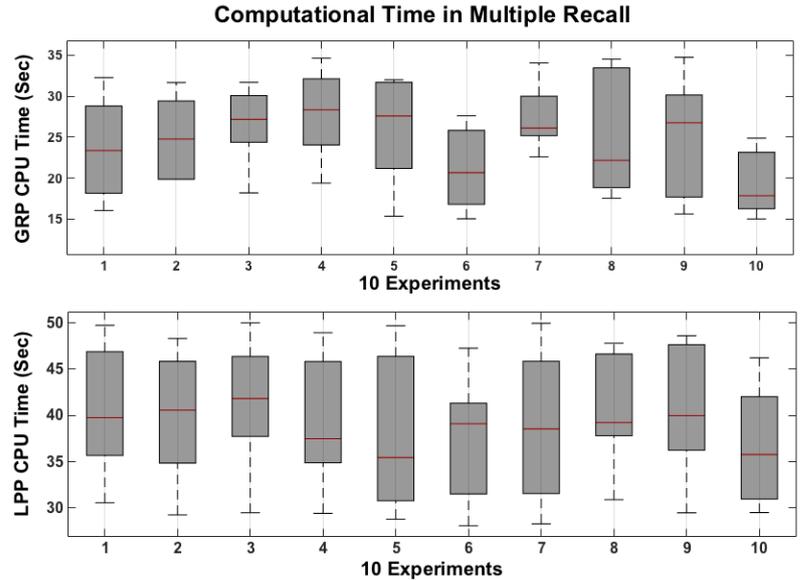
**Fig.19.** Stability of LPP and GRP computational time variation for different recall in 10 individual experiments

## 9 Conclusion

In this paper, a novel approach for enhancement of an underwater vehicle's autonomy for large-scale underwater mission was provided. This included a two-layer architecture, route planner in top level and path planner in low level, working interactively with each other and made vehicle capable of robust decision-making. Indeed, this research is an extension of previous study (M.Zadeh et al. 2016-d, 2016-e) in which the high and low level motion planner are designed in a separate modular format, so that the employed algorithms by each module can be easily replaced or upgraded. The main advantage of the proposed framework is having a modular and flexible structure that is compatible with a broad range of computational methods. The underwater mission, which conceptually is a kind of task assignment problem, was specified by accomplishing the maximum number of assigned tasks regarding the mission available time. By doing so, a series of diverse scenarios were designed to evaluate the performance and reliability of the proposed model. Simulation results showed that the proposed model is able to generate real-time near-optimal solutions that are relatively independent of both size and complexity of operation network. Therefore, the main objective of mission that was maximizing the mission productivity while keeping the vehicle safety was perfectly satisfied. Besides, the results indicated that the proposed model is good choice for operating in dynamic environment as it can excellently handle the influence of uncertainties through the mission.

As prospect for future research, we will plan to improve the level of vehicle's overall situation awareness by using the estimation of one step forward of mission operating filed changes and then feeding those to the model to generate the solutions for such a highly dynamic and uncertain missions. Besides, the functionality of the model will be investigated on a sea test trials. The modules will be upgraded with online replanning capability operating in a more realistic environment.

**Acknowledgment**
Somaiyeh M.Zadeh and Amirmehdi Yazdani are funded by Flinders International Postgraduate Research Scholarship (FIPRS) program, Flinders University of South Australia. This research is also supported through a FIPRS scheme from Flinders University. The authors declare that there is no conflict of interest.